\documentclass{article}

 \usepackage[preprint]{neurips_2025}

\usepackage[utf8]{inputenc} 
\usepackage[T1]{fontenc}    
\usepackage{hyperref}       
\usepackage{url}            
\usepackage{booktabs}       
\usepackage{amsfonts}       
\usepackage{nicefrac}       
\usepackage{microtype}      
\usepackage{xcolor}         
\usepackage{graphicx}
\usepackage{amsmath}
\usepackage{dsfont}

\usepackage{xspace}
\usepackage{amsmath} 
\usepackage{graphicx} 
\usepackage{booktabs}
\usepackage{multirow} 
\usepackage{algorithm} 
\usepackage{algpseudocode}
\usepackage{listings} 
\usepackage{wrapfig}
\usepackage{tabularx}
\usepackage[normalem]{ulem}
\useunder{\uline}{\ul}{}

\def\OURS{CLNet\xspace}

\title{\OURS: Cross-View Correspondence Makes a Stronger Geo-Localizationer}

\author{Xianwei Cao \And
Dou Quan \And
Shuang Wang \And
Ning Huyan \AND
Wei Wang \And
Yunan Li \And
Licheng Jiao \AND
\\
School of Artificial Intelligence, Xidian University \\
Xi'an, 710071, China
}

\begin{document}

\maketitle

\begin{abstract}
Image retrieval-based cross-view geo-localization (IRCVGL) aims to match images captured from significantly different viewpoints, such as satellite and street-level images. Existing methods predominantly rely on learning robust global representations or implicit feature alignment, which often fail to model explicit spatial correspondences crucial for accurate localization. In this work, we propose a novel correspondence-aware feature refinement framework, termed CLNet, that explicitly bridges the semantic and geometric gaps between different views. CLNet decomposes the view alignment process into three learnable and complementary modules: a Neural Correspondence Map (NCM) that spatially aligns cross-view features via latent correspondence fields; a Nonlinear Embedding Converter (NEC) that remaps features across perspectives using an MLP-based transformation; and a Global Feature Recalibration (GFR) module that reweights informative feature channels guided by learned spatial cues. The proposed CLNet can jointly capture both high-level semantics and fine-grained alignments. Extensive experiments on four public benchmarks, CVUSA, CVACT, VIGOR, and University-1652, demonstrate that our proposed CLNet achieves state-of-the-art performance while offering better interpretability and generalizability. 
\end{abstract}
\section{Introduction}
Image retrieval-based cross-view geo-localization (IRCVGL) obtains 
geographic location information of a given ground-view image by retrieving its corresponding satellite-view image from a database of geo-tagged satellite images \cite{hu2018cvmnet}. It is crucial for autonomous driving \cite{hu2023planningoriented}, augmented reality \cite{xiong2021augmenteda}, and robot localization  \cite{betke1997mobile}. 
The key challenge of IRCVGL lies in accurately matching two drastically different views corresponding to the same location.

Existing methods address this challenge from two main perspectives. They aim to reduce the perspective gap through geometric transformations~\cite{shi2019spatialaware, shi2020where, shi2022accurate} and view synthesis techniques~\cite{toker2021coming}. Additionally, existing methods focus on learning more robust and discriminative features by utilizing powerful backbones~\cite{yang2021crossview, zhu2022transgeo} or through contrastive learning with hard sample mining~\cite{deuser2023sample4geo}.

\begin{figure}[!htbp]
    \centering
    \includegraphics[width=1.\linewidth]{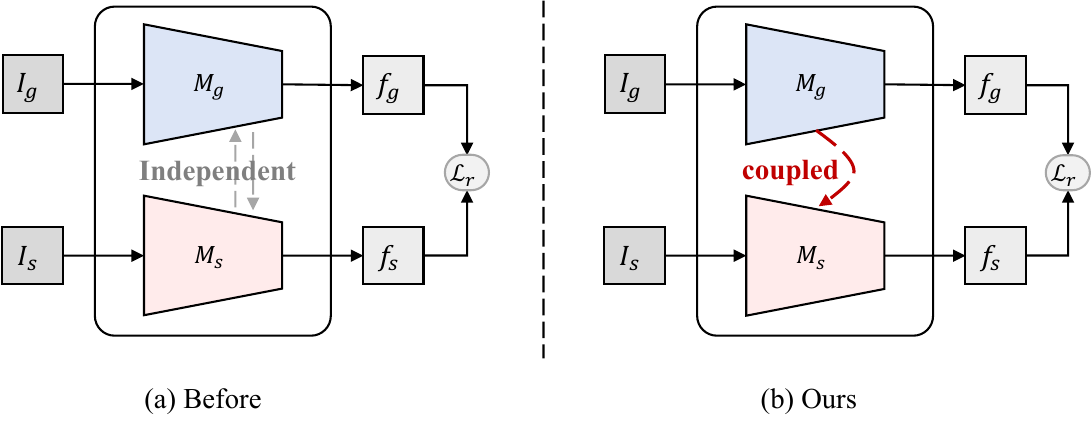}
    \caption{(a) previous independent learning streams and (b) our proposed correspondence learning network (CLNet), which couples two branches. This coupled learning approach enables interaction between ground-view and satellite-view streams to effectively learn relevant cross-view correspondence.}
    \vspace{-3mm}
    \label{fig:difference}
\end{figure}

\begin{figure*}[!ht]
  \centering
  \includegraphics[width=0.95\linewidth]{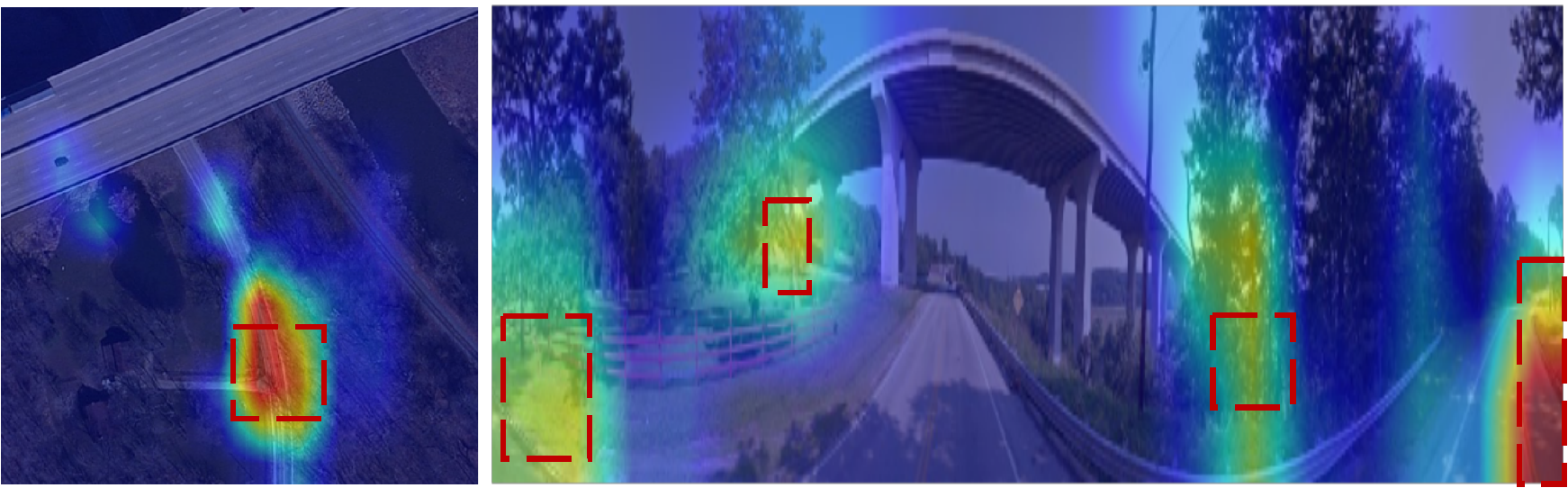}
  \caption{Grad-CAM \cite{selvaraju2017grad} visualization result of Sample4Geo \cite{deuser2023sample4geo} on the CVUSA dataset. As can be seen, the model always focuses on the correspondence information between the critical elements in the two views. Despite the huge difference in perspective, there is still rich corresponding information between the two views that can assist in improving the performance of IRCVGL.}
  \label{fig:motivation}
   \vspace{-3mm}
\end{figure*}

As illustrated in Fig.~\ref{fig:difference} (a), existing methods perform ground and satellite view feature learning independently, without any interaction across views. However, as shown in Fig.~\ref{fig:motivation}, we observe that a well-trained IRCVGL model mainly focuses on correspondences between the two views, similar to how humans match what they see with visual cues on a map. We argue that leveraging cross-view correspondences (e.g., roads, intersections, buildings) between two views is essential for accurate geo-localization. Existing independent feature learning methods fail to fully exploit cross-view correspondences. Therefore, their extracted feature representations merely rely on intra-view cues and are sensitive to viewpoint changes, resulting in suboptimal performance on IRCVGL.

While a straightforward solution is to enable feature-level interaction between views, naïve mechanisms typically incur prohibitive inference costs, $\mathcal{O}(n^2)$, rendering them impractical for large-scale deployment~\cite{zhang2020contextaware, wei2020multimodality}. This leads us to the central question: \textbf{How can we model cross-view correspondences effectively without sacrificing inference efficiency?}

To this end, we propose \textbf{CLNet} (\textbf{C}orrespondence \textbf{L}earning \textbf{Net}work), which explicitly models cross-view correspondences while maintaining efficient inference simultaneously. As illustrated in Fig.~\ref{fig:difference} (b), our core insight is to couple the ground and satellite learning streams through a learnable interaction mechanism that encodes both intra-view context and inter-view correspondences. Specifically, CLNet introduces ground-view neural maps to aggregate global intra-view contextual information. Secondly, we can generate the corresponding satellite-view neural map through a lightweight \textbf{N}eural bird's-\textbf{E}ye view \textbf{C}onverter (NEC), which captures latent correspondence mappings across views. To inject the learned correspondence information into the feature learning process, a \textbf{G}lobal-weighted \textbf{F}eature \textbf{R}efinement (GFR) module is designed to refine the extracted features under the guidance of neural maps, thereby enhancing consistency between views.

The main contributions of this work are summarized as follows:
\begin{itemize}
\item We propose \textbf{CLNet}, a novel approach that explicitly models cross-view correspondences for IRCVGL. By coupling multi-level feature learning of ground and satellite views, CLNet significantly improves cross-view geo-localization performance.
\item CLNet introduces three key modules, including a learnable view neural map, a Neural bird’s-Eye view Converter (NEC), and a Global Feature Recalibration (GFR) module, which can capture cross-view correspondences for enhancing cross-view consistency.
\item We conduct extensive experiments on four public benchmarks (e.g., CVUSA, CVACT, VIGOR, and University-1652), demonstrating that CLNet not only outperforms other methods in the IRCVGL but also exhibits strong generalization in cross-dataset and cross-region settings.
\end{itemize}

\section{Related Work}

The cross-view geo-localization task is broadly divided into two distinct sub-tasks: image retrieval-based geo-localization \cite{hu2018cvmnet, mithun2023cross} and 3-DoF pose estimation \cite{xia2023convolutional}. This study focuses on the former—image retrieval-based cross-view geo-localization (IRCVGL). Early approaches relied on hand-crafted descriptors to align ground-level and satellite images directly \cite{senlet2011framework, noda2011vehicle, lin2013cross, viswanathan2014vision, bansal2011geo, senlet2012satellite}, or attempted to reshape ground images into overhead, satellite-like formats to facilitate feature comparison \cite{senlet2011framework, viswanathan2014vision}. However, these methods struggled with the extreme perspective disparities between views.

With the rise of deep learning, robust feature representation learning methods are proposed for cross-view matching. \cite{workman2015location, workman2015wide} used off-the-shelf CNN features for cross-view matching, outperforming traditional descriptors. \cite{vo2016localizing} introduced Siamese and triplet networks with a soft-margin triplet loss to learn view-invariant embeddings. \cite{hu2018cvmnet} incorporated a Net-VLAD layer \cite{arandjelovic2016netvlad} into a Siamese framework and employed a weighted soft-margin ranking loss to accelerate training. Meanwhile, \cite{zhai2017predicting} utilized semantic segmentation to generate layout-based descriptors for satellite views.

To explicitly reduce the viewpoint gap, several methods focused on geometric alignment or view synthesis. \cite{shi2019spatialaware} used the polar coordinate transformation to warp satellite images into the ground-view perspective. \cite{toker2021coming} extended this by combining polar transformation with GAN-based image synthesis to generate a satellite-view image from its corresponding ground-view. \cite{shi2020optimal} introduced the Cross-View Feature Transformation (CVFT) module to explicitly map features between the two domains, improving alignment. Other works considered contextual cues. \cite{wang2021each} used square-ring partitioning to model the surrounding building context, \cite{liu2019lending} and \cite{shi2020where} integrated orientation cues to guide retrieval. More recently, \cite{zhu2021revisiting} and \cite{hu2022beyond} proposed fine-grained orientation estimation and activation-map-guided alignment to further improve performance in unaligned or partially aligned scenarios.

Despite these advances, most existing methods still process ground and satellite views independently, without explicit feature interaction between the two views. This limits their ability to model the rich geometric correspondences that are essential for robust geo-localization—particularly in cases of large viewpoint differences. The learned representations are often grounded solely in intra-view information, making cross-view alignment fragile and indirect. To address this issue, we introduce a structured mechanism for cross-view feature interaction, enabling the model to learn and exploit latent geometric relationships between views directly during training.

\section{Methods}

\begin{figure*}[!tbp]
    \centering
    \includegraphics[width=1.\linewidth]{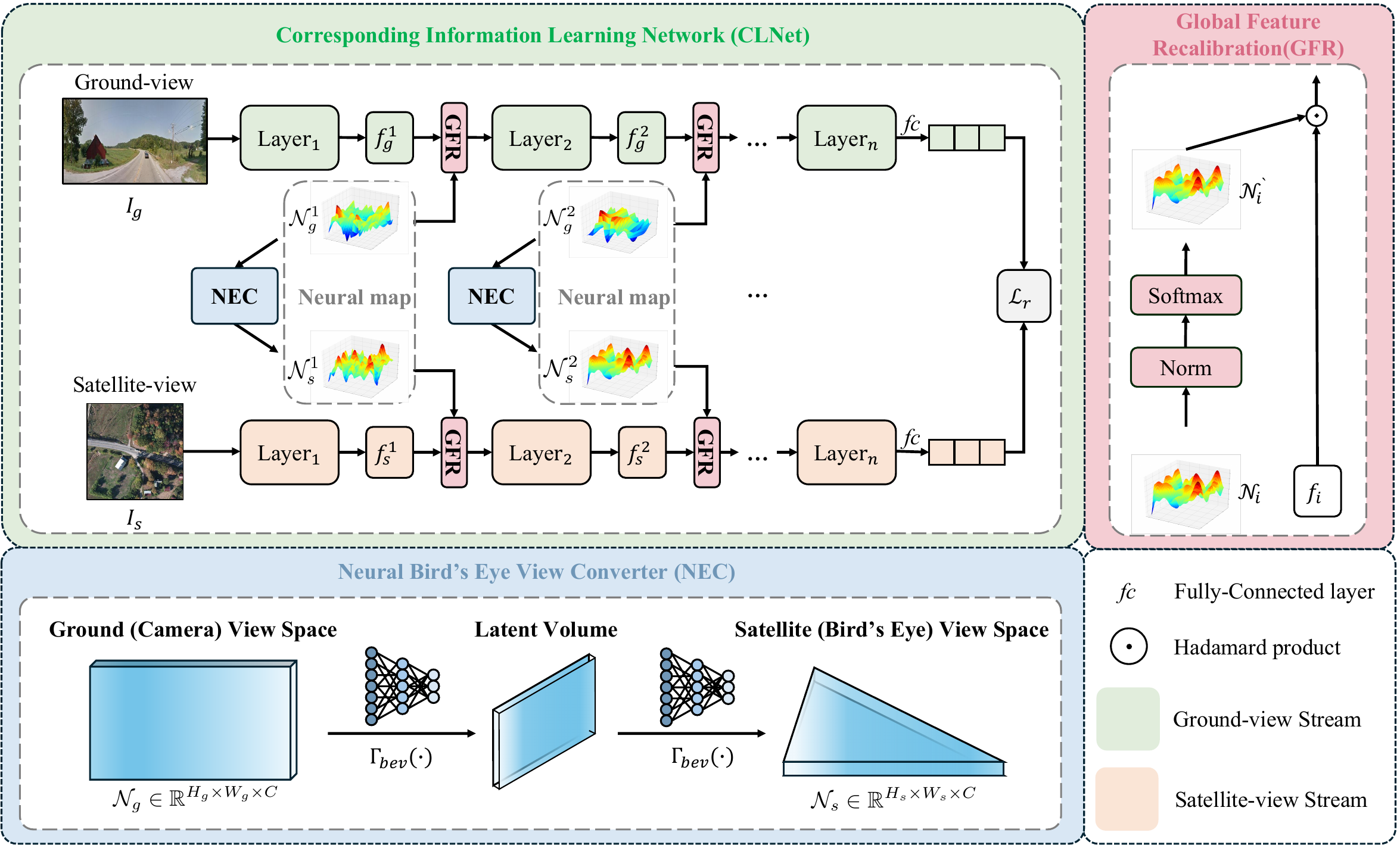}
    \caption{The overall pipeline of proposed \OURS. CLNet consists of three key components: (1) View neural maps; (2) Global Feature Recalibration (GFR); and (3) the Neural Bird’s Eye View Converter (NEC).}
    \label{fig:pipeline}
\end{figure*}

Given a ground-view image $I_g \in \mathbb{R}^{H_g \times W_g \times 3}$, the objective of IRCVGL is to retrieve its corresponding satellite-view image $I_s \in \mathbb{R}^{H_s \times W_s \times 3}$ from a geo-tagged database. The key challenge of IRCVGL lies in extracting discriminative and robust features from these two drastically different views. Following previous methods \cite{deuser2023sample4geo} and \cite{li2024unleashing}, we adopt ConvNeXt~\cite{liu2022convnet} as the backbone network for feature extraction. ConvNeXt consists of four sequential blocks, producing multi-scale feature representations for ground-view and satellite-view images, denoted as $f_g^i$ and $f_s^i$ ($i = 1, \dots, 4$), respectively.

A significant challenge in cross-view geo-localization arises from the substantial viewpoint discrepancy between ground-level and satellite images, which hampers the extraction of reliable correspondences. To enhance geo-localization performances, this paper tries to exploit correspondences across views and proposes a \textbf{C}orrespondence information \textbf{L}earning \textbf{Net}work (\OURS). Inspired by the concept of Inverse Perspective Mapping (IPM)~\cite{can2021structured, reiher2020sim2real}, CLNet aims to model the correspondence between two views and bridge the geometric gap between views in a learnable manner. 

As shown in Fig. \ref{fig:pipeline}, CLNet comprises three key modules, including view neural maps, Neural bird’s Eye view Converter (NEC), and a Global Feature Recalibration (GFR) module. Specifically, CLNet first introduces learnable view neural maps to aggregate global contextual information from the ground-view. Secondly, CLNet generates the corresponding satellite-view neural map through a lightweight NEC to enhance the view consistency. After that, CLNet employs a GFR module to refine the extracted features based on the neural maps of two views. The details of CLNet are described below.

\subsection{View Neural maps}
\label{sub-sec:view-neural-maps}

This paper captures the correspondence between two views according to the neural map and adopts their correspondence to enhance feature learning of the ground and satellite views. The view neural maps are a set of 3D learnable parameter matrices, which are defined as:
\begin{equation}
\mathcal{N}_v^i \in \mathbb{R}^{H^i_v \times W^i_v \times C^i},
\end{equation}
where $\mathcal{N}_v^i$ represents the neural map of the view $v$ in the $i_{th}$ feature layer, $v\in\{g, s\}$ represents the ground and satellite views. $\mathcal{N}_g^i$ aggregates the global context from the ground-view image. According to $\mathcal{N}_g^i$, we can acquire the corresponding satellite-view neural map, $\mathcal{N}_s^i$, through NEC. 

\begin{figure}[!ht]
    \centering
    \includegraphics[width=1.\linewidth]{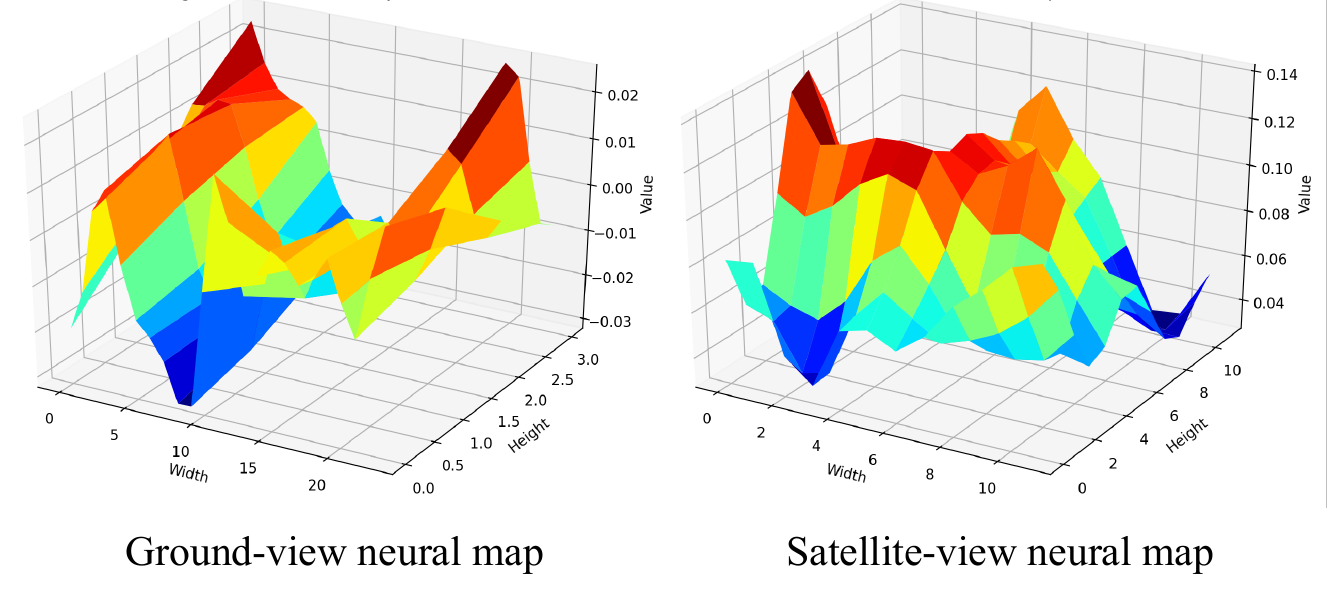}
    \caption{Visualization of the ground-view (left) and satellite-view (right) neural maps.}
    \label{fig:view_neural_map}
\end{figure}

Figure \ref{fig:view_neural_map} presents a pair of neural maps. The values in the neural map reflect the strength of the correspondence between the ground-view and satellite-view images. Higher (smaller) values indicate stronger (weaker) cross-view correspondences in local regions. 

\subsection{Neural Bird's Eye View Converter}
\label{sub-sec:NEC}

As shown at the bottom of Fig. \ref{fig:pipeline}, CLNet introduces the NEC module to generate the corresponding satellite-view neural map based on the ground-view neural map. Due to the significant geometric perspective difference between the ground-view and the satellite-view map, direct geometric coordinate conversion without prior knowledge of both camera poses can result in errors, blur, and even loss of information that could significantly assist in localization. Instead, we learn the geometric transformation from ground-view satellite-view space through a shallow neural network. This learning approach adaptively adjusts the important information required for matching the two viewpoints while suppressing interfering information. Specifically, NEC adopts a learnable view mapping function $\Gamma_{bev}$ that transforms the neural map of the ground-view into the satellite-view space. $\Gamma_{bev}$ can be implemented using a lightweight multi-layer perceptron (MLP):

\begin{equation}
\mathcal{N}_s = \Gamma_{bev}(\mathcal{N}_g),
\end{equation}
where $\mathcal{N}_g$ and $\mathcal{N}_s$ are ground-view and satellite-view neural maps, respectively. 

By projecting the neural map from the ground-view to the satellite-view, NEC implicitly models cross-view correspondences and effectively bridges the cross-view gap. Importantly, this operation introduces negligible computational overhead and facilitates shared reasoning between the two view-specific branches during inference. 

\subsection{Global Feature Recalibration}
\label{sub-sec:GFR}

To further enhance the feature consistency across views, this paper designs the \textbf{G}lobal \textbf{F}eature \textbf{R}ecalibration (GFR) module to recalibrate the extracted features of two views according to neural maps. The GFR module is formulated as follows: 
\begin{equation}\label{eq:gfr}
\begin{aligned}
{f_v^i}^{\prime} &= \operatorname{GFR}(f_v^i, \mathcal{N}_v^i) \\
&= f_v^i \odot \mathcal{N}_v^{\prime} + f_v^i, \quad v\in \{g,s\}, 
\end{aligned}
\end{equation}
where $\odot$ denotes the Hadamard product, and $\mathcal{N}_v^{\prime}$ is a globally normalized neural map. The residual connection in Eq.~\eqref{eq:gfr} allows the original features to be preserved while selectively emphasizing salient regions guided by the neural map.

\begin{equation}\label{eq:norm}
\mathcal{N}_v^{\prime} = \sigma\left(\frac{\mathcal{N}_v^i}{\|\mathcal{N}_v^i\|_2}\right),
\end{equation}
where $\sigma(\cdot)$ denoting the softmax operation applied over all spatial positions. The normalization and softmax operations ensure that $\mathcal{N}_v^{\prime}$ encodes a globally consistent importance distribution.

\subsection{Network optimization}
The whole framework of the proposed CLNet can be optimized by the infoNCE loss \cite{oord2018representation}. Supposing that the extracted features of the ground-view and satellite-view images are $f_g$ and $f_s$, respectively. The infoNCE loss can be represented as follows:
\begin{equation}
    \mathcal{L}_\text{infoNCE}(f_g, f_s) = -\log{
    \frac{ \exp{(f_g \cdot f_s^+) / \tau} }{ \sum_{j=0}^{M}{\exp{(f_g \cdot f_s^j) / \tau}} }},
\end{equation}
where $M$ denotes the number of satellite-view images in the database, $f_s^+$ is the matching satellite-view image of the $f_g$, and $\tau$ is a temperature parameter. The entire training framework of \OURS is outlined in Algorithm~\ref{alg:train_OURS}.   

\begin{algorithm}[H]
\caption{Training Procedure for \OURS}
\label{alg:train_OURS}
\begin{algorithmic}[1]
\State Initialize backbone $F_g$ and $F_s$ for ground and satellite view.
\State Initialize Neural Bird’s Eye View Converter $\Gamma_{bev}$.
\For{each feature level $i$}
\State Initialize ground-view neural map $\mathcal{N}_g^i$.
\State Obtain satellite-view neural map via $\Gamma_{bev}(\mathcal{N}_g^i)$.
\EndFor
\State \textbf{Input:} Ground-view and satellite-view image set $I_g$ and $I_s$.
\For{epoch = 1 to $N$}
\For{each ground-satellite image pair $x_g$ and $x_s$}
\For{each feature level $i$}
\State Extract features $f_g^i$ = $F_g^i(f_g^{i-1})$, $f_s^i$ = $F_s^i(f_s^{i-1})$.
\State Refine features using GFR:
\[
\begin{aligned}
f_g^i &= \operatorname{GFR}(f_g^i, \mathcal{N}_g^i), \\
f_s^i &= \operatorname{GFR}(f_s^i, \mathcal{N}s^i).
\end{aligned}
\]
\EndFor
\State Compute loss $\mathcal{L}_\text{infoNCE}(f_g, f_s)$.
\State Perform backpropagation and optimization.
\EndFor
\EndFor
\end{algorithmic}
\end{algorithm}

\section{Experiments}
\label{sec:experiments}

\begin{table*}[tbp]
\caption{Comparison results on CVUSA and CVACT datasets. * represents employing a polar transformation on satellite images. The best results are in bold.}
  \centering
  \resizebox{1.0\linewidth}{!}{
    \begin{tabular}{lcccc|cccc|cccc}
    \toprule
    \multirow{2}[2]{*}{Approach} & \multicolumn{4}{c}{CVUSA} & \multicolumn{4}{c}{CVACT\_val} & \multicolumn{4}{c}{CVACT\_test} \\
\cmidrule{2-13}          & R@1 & R@5 & R@10 & R@1\% & R@1 & R@5 & R@10 & R@1\% & R@1 & R@5 & R@10 & R@1\% \\
    \midrule
    SAFA* (NeurIPS'2019) \cite{shi2019spatialaware}      & 89.84  & 96.93  & 98.14  & 99.64  & 81.03  & 92.80  & 94.84  & 98.17  & 55.50  & 79.94  & 85.08  & 94.49  \\
    CDE*  (CVPR'2021) \cite{toker2021coming}       & 92.56  & 97.55  & 98.33  & 99.57  & 83.28  & 93.57  & 95.42  & 98.22  & 61.29  & 85.13  & 89.14  & 98.32  \\
    L2LTR  (NeurIPS'2021) \cite{yang2021crossview}       & 91.99  & 97.68  & 98.65  & 99.75  & 83.14  & 93.84  & 95.51  & 98.40  & 58.33  & 84.23  & 88.60  & 95.83  \\
    L2LTR*     & 94.05  & 98.27  & 98.99  & 99.67  & 84.89  & 94.59  & 95.96  & 98.37  & 60.72  & 85.85  & 89.88  & 96.12  \\
    TransGeo  (CVPR'2022) \cite{zhu2022transgeo}    & 94.08  & 98.36  & 99.04  & 99.77  & 84.95  & 94.14  & 95.78  & 98.37  & -     & -     & -     & - \\
    GeoDTR  (AAAI'2023) \cite{zhang2023cross}      & 93.76  & 98.47  & 99.22  & 99.85  & 85.43  & 94.81  & 96.11  & 98.26  & 62.96  & 87.35  & 90.70  & 98.61  \\
    GeoDTR*    & 95.43  & 98.86  & 99.34  & 99.86  & 86.21  & 95.44  & 96.72  & \textbf{98.77} & 64.52  & 88.59  & 91.96  & \textbf{98.74 } \\
    SAIG-D  (arXiv'2023) \cite{zhu2023simple} & 96.34  & 99.10  & 99.50  & 99.86 & 89.21  & 96.07  & 97.04  & 98.74  & 67.49  & 89.39  & 92.30  & 96.80  \\
    Sample4Geo  (ICCV'2023) \cite{deuser2023sample4geo}  & 98.68  & 99.68  & \textbf{99.78}  & 99.87  & 90.81 & \textbf{96.74} & \textbf{97.48} & \textbf{98.77}  & 71.51 & 92.42 & 94.45 & 98.70 \\
    GeoDTR+  (TPAMI'2024) \cite{zhang2024geodtr+}     & 95.40  & 98.44  & 99.05  & 99.75  & 87.61  & 95.48  & 96.52  & 98.34  & 67.57  & 89.84  & 92.57  & 98.54  \\
    Ours  & \textbf{98.77} & \textbf{99.71} & 99.77 & \textbf{99.89} & \textbf{91.22} & 96.61 & 97.38 & 98.63 & \textbf{71.89} & \textbf{92.65} & \textbf{94.55} & \textbf{98.65} \\
    \bottomrule
    \end{tabular}%
    }
  \label{tab:res_CVUSA_CVACT}%
\end{table*}%

\subsection{Datasets}
 We evaluate CLNet using four benchmark datasets CVUSA \cite{workman2015wide}, CVACT \cite{liu2019lending}, VIGOR \cite{zhu2021vigor}, and University-1652 \cite{zheng2020university}. CVUSA and CVACT contain aligned ground-satellite image pairs from North America and Australia, respectively, with large training sets and a one-to-one correspondence setup. VIGOR introduces a more challenging setting with semi-positive samples and cross-region generalization. University-1652 focuses on drone-to-satellite matching with many-to-one relationships, which require models to generalize across complex viewpoints. 

\subsection{Datasets}
\noindent{\textbf{CVUSA}~\cite{workman2015wide}} contains 35,532 pairs of images for training and 8,884 pairs of images for evaluation. The street view images are $224 \times 1232$, and the size of the satellite images is $750 \times 750$. By warping the ground-view images through the camera's extrinsic parameters, all ground-satellite image pairs are aligned, and all ground-view images are located in the center of the satellite images. 

\noindent{\textbf{CVACT}~\cite{liu2019lending}} is similar to CVUSA overall, except that it contains an additional test set, totaling 92,802 image pairs. In addition, most of the scenes in the dataset are urban from Australia, which can be considered as a different scene style compared with the CVUSA dataset. At the same time, it has a larger image resolution. The size of the ground images is $832 \times 1664$, while the size of the satellite images is $1200 \times 1200$. The query images and reference images in CVUSA and CVACT have a one-to-one relationship in that each query corresponds to one reference, and the ground images and satellite images are also center-aligned.

\noindent{\textbf{VIGOR}~\cite{zhu2021vigor}} was collected from four cities in the United States, \textit{i.e.}, New York, Seattle, San Francisco, and Chicago. It contains two partitions, which are Same-area and Cross-area. The Same-area uses data from all cities for training and evaluation. The Cross-area uses New York and Seattle as the training set and San Francisco and Chicago for evaluation. Unlike the two datasets mentioned above, the positional relationship between the ground view image and the satellite image in this dataset contains two situations: positive and semi-positive. Specifically, each ground view image contains three semi-positive satellite images, which means the ground view image is located in the non-center region of the three satellite images. Such a setting leads to a more difficult situation.

\noindent{\textbf{University-1652}~\cite{zheng2020university}} is proposed to bridge the gap between the drone-view and satellite-view. Different from the previous three datasets that contain ground-view and satellite-view image pairs, the University-1652 dataset includes multiple drone-view images for the same building. There are 50,218 images for training, and 54 drone-view images for each satellite-view image. For the benchmarking part, over 1,652 buildings of 72 universities are captured to ensure the generalization of the test. Additionally, since not all satellite-view images have their corresponding reference drone-view images, we report the average precision (AP) for this dataset.

\subsection{Implementation Details}
The two-view learning branches are based on ConvNeXT-B \cite{liu2022convnet}. For CVUSA and CVACT, the satellite images and the ground images are resized to $384 \times 384$ and $140 \times 768$, respectively. For the VIGOR dataset, they are resized to $384 \times 384$  and $384 \times 768$. For University-1652, we resize all view images from $512\times 512$ to $384\times 384$. Only horizontal flipping and rotation are used for satellite images and synchronous shift for ground images \cite{zhang2023cross, deuser2023sample4geo}. For a fair comparison, the rest of the experimental settings follow the same content as Sample4Geo \cite{deuser2023sample4geo}. 

In all our experiments, the batch size is 128 for CVUSA and CVACT, and 96 for VIGOR, We train the model on 8 $\times$ NVIDIA RTX 3090 GPUs and use 16 worker threads for PyTorch dataloader. We also employ random flip, random rotate, and ColorJitter for data augmentation. In addition, the optimizer is AdamW, with a learning rate of 0.001, and the other config is the default following PyTorch. We also set the training epoch to 40 and introduce cosine decay with a warmup as the learning rate schedule for better convergence.

\subsection{Evaluation Metrics}
We evaluate the performance of our method using \textit{Recall at top-k} (Recall@k), a widely adopted metric in cross-view geo-localization~\cite{vo2016localizing, workman2015location}. Recall@k measures the percentage of query images whose correct match appears within the top-$k$ retrieved results. For the VIGOR dataset, we additionally report the \textit{Hit Rate}~\cite{zhu2021vigor}, which considers whether the top-1 retrieved result belongs to the set of positive or semi-positive matches. It is defined as:
\begin{equation}
   \text{Recall@k} = \frac{1}{N} \sum_{i=1}^{N} \mathbb{I}\left[\text{rank}(q_i, k_i^+) \leq k \right],
\end{equation}
where $N$ is the number of queries, $q_i$ is the $i$-th query, $k_i^+$ is the corresponding ground-truth reference, $\text{rank}(q_i, k_i^+)$ denotes the rank of the correct match for $q_i$ in the retrieval list, and $\mathbb{I}[\cdot]$ is the indicator function.

For the VIGOR dataset, we additionally report the \textit{Hit Rate}~\cite{zhu2021vigor}, which considers whether the top-1 retrieved result belongs to the set of positive or semi-positive matches.
\begin{equation}
   \text{Hit Rate} = \frac{1}{N} \sum_{i=1}^{N} \mathbb{I}\left[r_1^i \in \mathcal{P}_i \right],
\end{equation}
where $r_1^i$ is the top-1 retrieved result for query $q_i$, and $\mathcal{P}_i$ is the set of valid (positive and semi-positive) references. The metric reflects the model's ability to retrieve any valid match in the first position under partial alignment settings.

These two metrics jointly reflect the retrieval accuracy and robustness of the method under standard and partially aligned settings.
 
\begin{table*}[!ht]
\footnotesize
\caption{Cross-dataset evaluation results, \textit{i.e.}, trained on CVUSA and evaluated on CVACT, and vice versa.  * represents employing a polar transformation on satellite images. The best results are in bold.}
  \centering
  \resizebox{1.\linewidth}{!}{
    \begin{tabular}{lcccc|cccc|c}
    \toprule
    \multirow{2}[2]{*}{Approach} & \multicolumn{4}{c|}{CVUSA $\rightarrow$ CVACT} & \multicolumn{4}{c}{CVACT $\rightarrow$ CVUSA} & \multirow{2}[2]{*}{Average}\\
\cmidrule{2-9}          & \multicolumn{1}{l}{R@1} & \multicolumn{1}{l}{R@5} & \multicolumn{1}{l}{R@10} & \multicolumn{1}{l|}{R@1\%} & \multicolumn{1}{l}{R@1} & \multicolumn{1}{l}{R@5} & \multicolumn{1}{l}{R@10} & \multicolumn{1}{l}{R@1\%} \\
    \midrule
    DSM* (CVPR'2020) \cite{shi2020where}               & 33.66 & 52.17 & 59.74 & 79.67  & 18.47  & 34.46  & 42.28  & 69.01 & 48.68  \\
    L2LTR* (NeurIPS'2021) \cite{yang2021crossview}     & 47.55 & 70.58 & 77.39 & 91.39  & 33.00  & 51.87  & 60.63  & 84.79 & 64.65  \\
    TransGeo (CVPR'2022) \cite{zhu2022transgeo}        & 37.81 & 61.57 & 69.86 & 89.14  & 18.99  & 38.24  & 46.91  & 88.94 & 56.43  \\
    GeoDTR* (AAAI'2023) \cite{zhang2023cross}          & 53.16 & 75.62 & 81.90 & 93.80  & 44.07  & 64.66 & 72.08  & 90.09  & 71.92  \\
    Sample4Geo (ICCV'2023) \cite{deuser2023sample4geo} & 56.62 & 77.79 & \textbf{87.02}  & 94.69  & \textbf{44.95} & \textbf{64.36} & 72.10  & 90.65 & 73.52 \\
    \OURS (Ours)                                       & \textbf{59.17} & \textbf{78.67} & 85.28 & \textbf{94.78} & 43.05 & 63.95 & \textbf{73.37} & \textbf{90.71} & \textbf{73.63} \\ 
    \bottomrule
    \end{tabular}%
    }
  \label{tab:CVUSA-CVACT}%
\end{table*}%

\subsection{Comparison Results}
We evaluate the effectiveness of our method on four public datasets: CVUSA, CVACT, VIGOR, and University-1652. To clearly demonstrate the advantages of our approach, we divide the experiments into three categories: (1) comparisons with state-of-the-art models on the same scenes, (2) cross-scene evaluations, and (3) generalization tests in more challenging scenarios. Specifically, \textit{CVUSA}, \textit{CVACT}, and \textit{VIGOR-Same} serve as standard benchmarks for evaluating retrieval performance under aligned ground-satellite image pairs. The \textit{CVUSA $\rightarrow$ CVACT}, \textit{CVACT $\rightarrow$ CVUSA}, and \textit{VIGOR-Cross} settings introduce additional challenges, such as off-center matching and cross-region generalization, and are used to assess model robustness. \textit{University-1652} further extends the evaluation to cross-domain and cross-altitude scenarios, where the Drone-to-Satellite and Satellite-to-Drone tasks reflect the model’s ability to generalize beyond typical ground-level viewpoints.

We compare our method with recent state-of-the-art approaches, including TransGeo~\cite{zhu2022transgeo}, GeoDTR+~\cite{zhang2024geodtr+}, and Sample4Geo~\cite{deuser2023sample4geo}, and report quantitative results across all settings. Benefiting from our proposed cross-view correspondence learning strategy, CLNet consistently outperforms existing methods and sets new state-of-the-art performance on multiple benchmarks.

\noindent{\textbf{The performance on the same scenes.}} We report the quantitative results of the proposed \OURS on the CVUSA and CVACT datasets in Table \ref{tab:res_CVUSA_CVACT}. On the CVUSA dataset, our \OURS achieves an R@1 score of 98.77\%, outperforming the prior SOTA method, Sample4Geo \cite{deuser2023sample4geo}, by a margin of 0.09\%. For R@5, our method also sets a new benchmark at 99.71\%, exceeding the prior best by 0.03\%. Although the R@10 score of 99.77\% is marginally lower by 0.01\% compared to the highest score, it remains comparable and well-aligned with the top-performing methods. Notably, the R@1\% metric sees a modest improvement of 0.02\%, further affirming the robustness of our approach. 

On the CVACT dataset, our \OURS continues to demonstrate its competitive edge. For the validation set, it achieves results on par with prior methods, with an R@1 of 96.61\% and R@1\% of 98.63\%. On the test set, our approach delivers an R@1 score of 71.89\%, a substantial improvement of 0.38\% over the best competing method. Additionally, the R@5 and R@10 metrics improve by 0.23\% and 0.10\%, respectively. Importantly, these achievements are obtained without employing polar transformations for satellite images or any other technical tricks, only introduce cross-view corresponding information. These results highlight the importance of incorporating correspondence information in IRCVGL.

To further evaluate the model’s cross-region generalization ability, we conduct experiments under two distinct settings on the VIGOR dataset: Same-area and Cross-area. In the Same-area setting, both the training and validation data are collected from the same geographic regions, whereas in the Cross-area setting, the training and test sets come from different, non-overlapping regions. As reported in Table \ref{tab:res_VIGOR}, the experimental results demonstrate the effectiveness of our proposed method. In the Same-area setting, CLNet achieves state-of-the-art performance, outperforming or closely matching the previous best method, Sample4Geo. Specifically, CLNet achieves the highest performance with a Recall@1 of 77.96\%, Recall@5 of 95.75\%, and Recall@10 of 97.28\%. Additionally, our approach obtains the highest Hit Rate at 90.09\%, underscoring its robustness in accurately recognizing spatial correspondences within the same geographic region.

\begin{table}[tbp]
\caption{Quantitative results on University-1652 dataset. The best results are indicated in bold.}
  \centering
    \begin{tabular}{lcccc}
    \toprule
    \multirow{2}[2]{*}{Approach} & \multicolumn{2}{c}{Drone2Sat} & \multicolumn{2}{c}{Sat2Drone}\\
          & R@1   & AP    & R@1   & AP\\
    \midrule
    LPN~\cite{wang2021each}   & 75.93 & 79.14 & 86.45 & 74.79\\
    SAIG-D~\cite{zhu2023simple} & 78.85 & 81.62 & 86.45 & 78.48 \\
    DWDR~\cite{wang2024learning}  & 86.41 & 88.41 & 91.3  & 86.02 \\
    MBF~\cite{zhu2023uav}   & 89.05 & 90.61 & 92.15 & 84.45 \\
    Sample4Geo~\cite{deuser2023sample4geo} & 92.65 & 93.81 & 95.14 & 91.39 \\
    Ours  & \textbf{93.02} & \textbf{94.18} & \textbf{98.31} & \textbf{91.63} \\
    \bottomrule
    \end{tabular}%
  \label{tab:res_university}%
\end{table}%

\noindent{\textbf{The performance on cross scenes.}}  To assess the generalization ability of our proposed \OURS, we conducted cross-dataset validation experiments by training on CVUSA and evaluating on CVACT, and vice versa (CVUSA $\rightarrow$ CVACT \& CVACT $\rightarrow$ CVUSA). The quantitative results are reported in Table \ref{tab:CVUSA-CVACT}. Overall, \OURS achieves best performances on average Recall, which is 73.63\%. For the CVUSA $\rightarrow$ CVACT setting, our \OURS demonstrates strong performance, achieving an R@1 score of 59.17\%, which surpasses the previous best approach, Sample4Geo \cite{deuser2023sample4geo}, by a margin of 2.55\%. Furthermore, our model attains competitive scores for R@5 (78.67\%) and R@1\% (94.78\%), with improvements of 0.88\% and 0.09\%, respectively, compared to the state-of-the-art. Although the R@10 score slightly declines by 2.74\%, the overall results still indicate that \OURS effectively generalizes to unseen data when trained on CVUSA and tested on CVACT. On the other hand, for the CVACT $\rightarrow$ CVUSA setting, the performance of \OURS is relatively less pronounced. While it achieves a competitive R@10 score of 73.37\% (a 1.27\% improvement) and an R@1\% score of 90.71\% (a 0.06\% improvement), the R@1 metric drops to 43.05\%, which is 1.90\% lower than the previous best method, GeoDTR* \cite{zhang2023cross}. This discrepancy highlights the challenges of transferring knowledge learned on CVACT to CVUSA. However, in terms of average recall, our method achieves the best performance compared to other methods, indicating that, overall, the introduction of correspondence information is effective.

Under the more challenging VIGOR-Cross setting, our method remains highly competitive, achieving a Recall@1 of 61.20\%, closely following Sample4Geo, the current top-performing method (61.70\%). Our results show strong generalization capabilities, outperforming other advanced approaches such as TransGeo, GeoDTR, GeoDTR+, and SAIG-D, particularly evident in the Recall@5 (83.45\%) and Recall@10 (87.94\%) metrics. The Hit Rate also remains robust at 69.55\%, further affirming the method's generalization capability. We also reproduce the baseline results for fair comparison. These results highlight the efficacy of explicitly modeling cross-view correspondence through our novel architecture.

\begin{table*}[!tbp]
\footnotesize
  \centering
  \caption{Quantitative results on VIGOR dataset, where * represent the results we have reproduced.. The best results are indicated in bold.}
  \resizebox{1.\linewidth}{!}{
    \begin{tabular}{lccccc|ccccc}
    \toprule
    \multirow{2}[1]{*}{Approach} & \multicolumn{5}{c}{Same-area} & \multicolumn{5}{c}{Cross-area} \\
\cline{2-11} & \multirow{1}[1]{*}{R@1}   & \multirow{1}[1]{*}{R@5} & \multirow{1}[1]{*}{R@10} & \multirow{1}[1]{*}{R@1\%} & \multirow{1}[1]{*}{Hit Rate} & \multirow{1}[1]{*}{R@1} & \multirow{1}[1]{*}{R@5} & \multirow{1}[1]{*}{R@10} & \multirow{1}[1]{*}{R@1\%} & \multirow{1}[1]{*}{Hit Rate} \\
    \midrule
    VIGOR (CVPR'2021) \cite{zhu2021vigor} & 41.07 & 65.81 & 74.05 & 98.37 & 44.71 & 11.00 & 23.56 & 30.76 & 80.22 & 11.64 \\
    TransGeo (CVPR'2022) \cite{zhu2022transgeo}  & 61.48 & 87.54 & 91.88 & 99.56 & 73.09 & 18.99 & 38.24 & 46.91 & 88.94 & 21.21 \\
    GeoDTR (AAAI'2023) \cite{zhang2023cross} & 56.51 & 80.37 & 86.21 & 99.25 & 61.76 & 30.02 & 52.67 & 61.45 & 94.4  & 30.19 \\
    SAIG-D (arXiv'2023) \cite{zhu2023simple} & 65.23 & 88.08 &   -   & \textbf{99.68} & 74.11 & 33.05 & 55.94 &   -   & 94.64 & 36.71 \\
    Sample4Geo (ICCV2023) \cite{deuser2023sample4geo} & 77.86 & 95.66 & 97.21 & 99.61 & 89.82 & \textbf{61.70} & \textbf{83.50} & \textbf{88.00} & \textbf{98.17} & \textbf{69.87} \\
    GeoDTR+ (TPAMI'2024) \cite{zhang2024geodtr+} & 59.01 & 81.77 & 87.1  & 99.07 & 67.41 & 36.01 & 59.06 & 67.22 & 94.95 & 39.40 \\
    Sample4Geo * & 77.47 & 95.29 & 96.87 & 99.61 & 89.12 & 60.99 & 83.09 & 87.63 & 98.19 & 69.42 \\
    \OURS (Ours) & \textbf{77.96} & \textbf{95.75} & \textbf{97.28} & 99.67 & \textbf{90.09} & 61.20 & 83.45 & 87.94 & \textbf{98.17} & 69.55 \\
    \bottomrule
    \end{tabular}%
    }
  \label{tab:res_VIGOR}%
\end{table*}%

\begin{figure*}[!tbp]
\footnotesize
    \centering
    \includegraphics[width=1.0\linewidth]{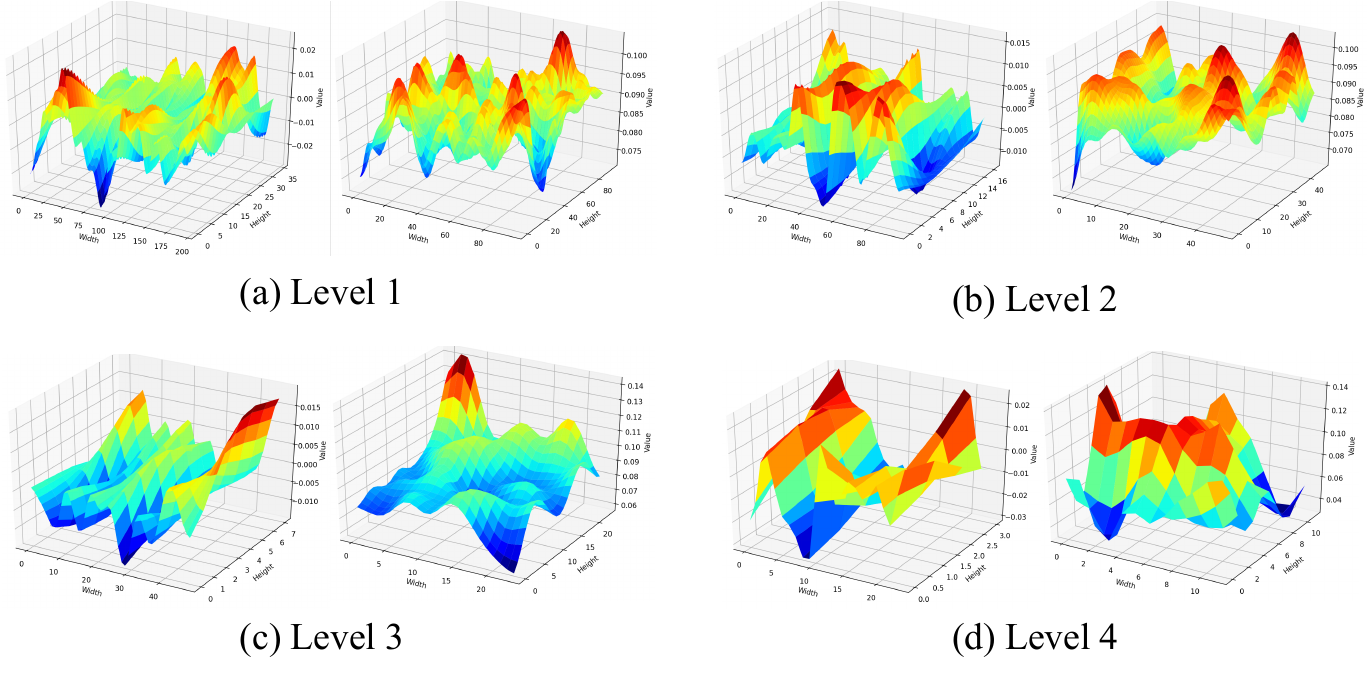}
    \caption{Visualization of view neural maps from ground-view (left in each pair) and satellite-view (right in each pair) across four feature levels (Levels 1–4, arranged from top to bottom and left to right). For better interpretability, Gaussian smoothing (kernel sizes: 5–1 from Level 1–4) is applied to highlight response patterns. Zoom in for details.    
    }
    \label{fig:visualization}
\end{figure*}

\noindent{\textbf{The generalization test on University-1652.}}
To evaluate cross domain and cross altitude scenarios, we perform and report our quantitative results in Table \ref{tab:res_university} for the Drone2Sat and Sat2Drone tasks. For Drone2Sat task, our method achieves an R@1 of 92.60\%, which is on par with the state-of-the-art Sample4Geo (92.65\%). In terms of AP, our approach records 93.75\%, slightly lower than Sample4Geo (93.81\%) but significantly outperforming other competing methods such as MBF (90.61\%) and DWDR (88.41\%). These results demonstrate that our method effectively retrieves the corresponding satellite images given drone-view queries, leveraging global intra-view and cross-view correlational information. The marginal difference in AP suggests potential improvements in handling edge cases and adversarial classes. In the Sat2Drone task, our method achieves an R@1 of 95.15\%, surpassing Sample4Geo (95.14\%) and setting a new state-of-the-art. Additionally, our method achieves an AP of 91.63\%, also exceeding Sample4Geo (91.39\%) and other competing methods. These results highlight the robustness of our approach in generalizing to diverse drone perspectives and effectively addressing the challenges posed by the adversarial test set. The consistent improvements in both R@1 and AP metrics validate the efficacy of our modules in capturing critical cross-view relationships. The results on the University-1652 dataset confirm the adaptability and generalization capacity of our approach across different domains, including 1:n mapping scenarios. The slight improvements over Sample4Geo in the Sat2Drone task further demonstrate the effectiveness of our method in addressing the unique challenges posed by diverse and adversarial datasets. Future work could explore strategies to bridge the small performance gap in AP for the Drone2Sat task while maintaining the overall robustness and efficiency of the model.

\subsection{Ablation Study}
\label{sub-sec:ablation}

To understand the effectiveness of our core idea, correspondence information learning, we conduct ablation studies on each component of \OURS. Due to its manageable scale and clear validation setup, we use the University-1652 dataset for all ablation experiments. The results are summarized in Table~\ref{tab:ablation-component}, with each configuration denoted as \#1–\#6. Our proposed \OURS framework comprises three key components: view neural maps, the \textbf{G}lobal \textbf{F}eature \textbf{R}ecalibration (GFR) module, and the \textbf{N}eural bird’s-\textbf{E}ye view \textbf{C}onverter (NEC). We perform controlled experiments to isolate the contribution of each module. In addition, we investigate the effect of residual connections within the GFR module, specifically analyzing the influence of the view neural maps ($\mathcal{N}^i$) and the extracted feature maps ($f^i$). These evaluations provide insight into how each component contributes to overall performance.



\begin{table}[!tbp]
\caption{Ablation study of CLNet on the University-1652 dataset. GFR and NEC denote the Global Feature Recalibration module and Neural Bird’s-Eye View Converter, respectively. $\mathcal{N}^i$ and $f^i$ indicate the use of view neural maps and feature maps within GFR. Norm refers to whether $\ell_2$ normalization is applied. Performance is reported in terms of Recall@1.
}
  \centering
    \begin{tabular}{ccccccc}
    \toprule
    Num.  &          GFR     &    Norm    & GFR ($\mathcal{N}^i$) & GFR ($f^i$) & NEC        & R@1 \\
    \midrule
    \#1    &  \checkmark       &            &                       &             &            & 91.58 \\
    \#2    &\checkmark       &            &            &                       & \checkmark & 92.10 \\
    \#3    &\checkmark      &            & \checkmark &                       & \checkmark & 92.65 \\
    \#4    &\checkmark      &            &            & \checkmark            & \checkmark & 92.42 \\
    \#5    &\checkmark        & \checkmark & \checkmark &                       & \checkmark & 93.02 \\
    \#6    &\checkmark         & \checkmark &            & \checkmark            & \checkmark & 92.75 \\
    \bottomrule
    \end{tabular}%
  \label{tab:ablation-component}%
\end{table}%

\begin{table}[!tbp]
  \centering
  \caption{Comparison of computational complexity and inference speed between CLNet and Sample4Geo, measured in FLOPs (G), MACs (G), and inference time (ms).} 
    \begin{tabular}{lccc}
    \toprule
    Methods & \multicolumn{1}{l}{flops(G)} & \multicolumn{1}{l}{macs(G)} & \multicolumn{1}{l}{Time(ms)} \\
    \midrule
    Sample4Geo & 2122.7 & 1059.4 & 315.9 \\
    Ours  & 676.1 & 337.4 & 341.4 \\
    \bottomrule
    \end{tabular}%
  \label{tab:params}%
\end{table}%

\noindent{\textbf{Effectiveness of Neural Bird's Eye View Converter (NEC).}} 
The baseline result, using only two sets of view neural maps, achieves 91.58\% R@1 (Config \#1). Incorporating NEC alone improves performance to 92.10\% (Config \#2), demonstrating the importance of modeling cross-view correspondences through a learned mapping.

\noindent{\textbf{Effectiveness of Residual design in \textbf{G}lobal \textbf{F}eature \textbf{R}ecalibration (GFR) module.}} 
When NEC is combined with view neural maps $\mathcal{N}^i$, performance further improves to 92.65\% (Config \#3), suggesting that integrating global intra-view information via residual connections in GFR benefits the feature refinement process. Alternatively, incorporating residual connections from the feature maps $f^i$ yields 92.42\% (Config \#4), indicating that while local features are beneficial, the global priors provided by neural maps offer stronger guidance for feature recalibration.

\noindent\textbf{Impact of $\ell_2$-Based Feature Scaling.}  
The best performance of 93.02\% R@1 is achieved when both $\ell_2$ normalization and view neural maps are used within GFR (Config \#5). This result highlights that combining feature scaling with global feature recalibration provides an effective strategy for integrating correspondence information into the learning process.

\noindent\textbf{Computational Complexity and Efficiency Analysis.}
To quantitatively evaluate the computational efficiency of our proposed CLNet, we compare its inference complexity and runtime against the recent state-of-the-art method, Sample4Geo \cite{deuser2023sample4geo}. As reported in Table~\ref{tab:params}, CLNet significantly reduces computational complexity, requiring approximately 68.2\% fewer FLOPs (from 2122.7G to 676.14G) and 68.1\% fewer MACs (from 1059.4G to 337.421G). Notably, despite substantial computational savings, CLNet maintains comparable inference speed to Sample4Geo (341.36 ms vs. 315.87 ms per query). This result demonstrates that our correspondence-aware approach achieves superior retrieval performance with considerably lower computational overhead.

\subsection{Visualization}

\begin{figure*}[!tb]
    \centering
    \includegraphics[width=1.\linewidth]{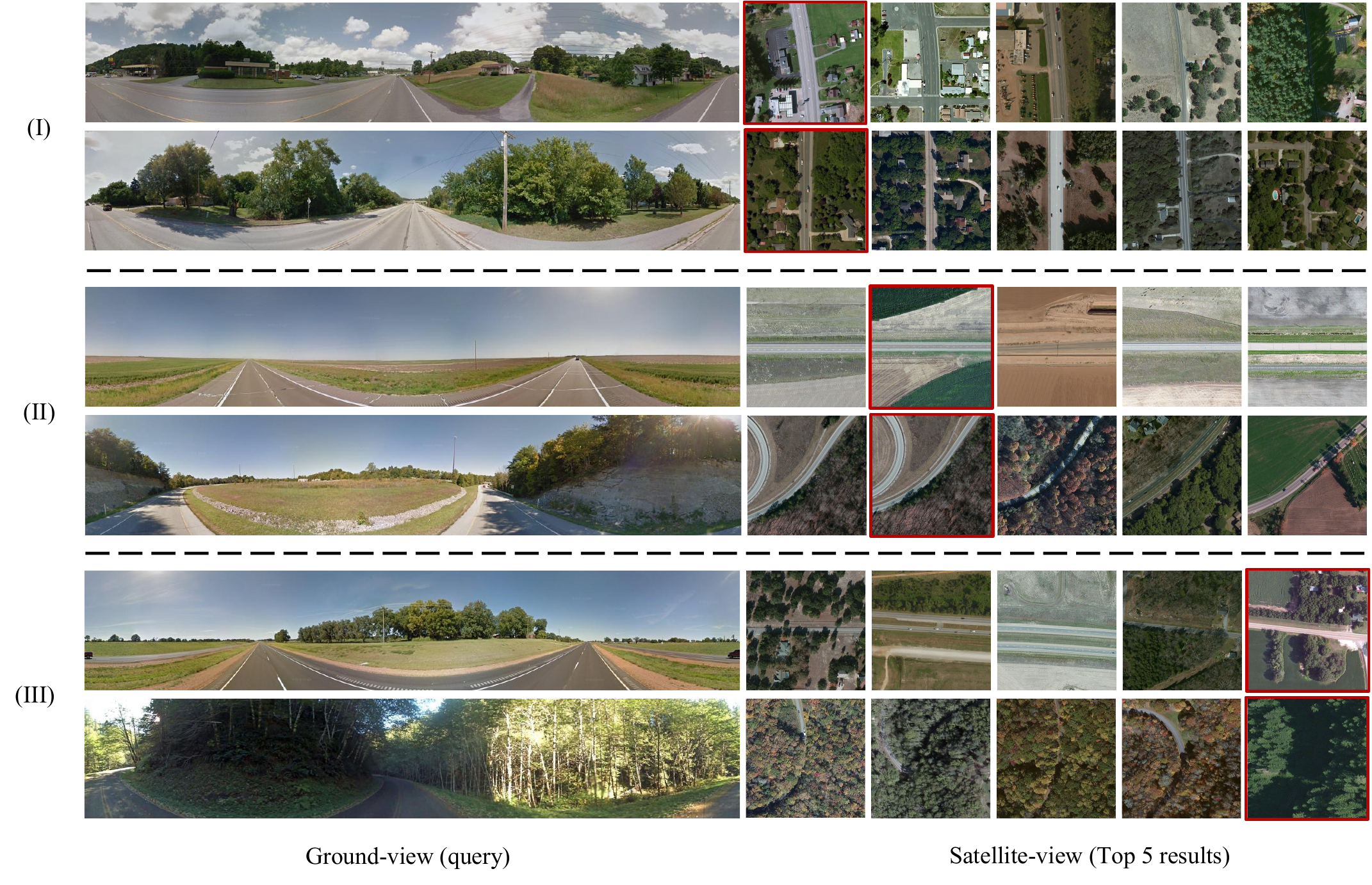}
    \caption{Qualitative retrieval results of our CLNet on the CVUSA dataset. Left: ground-view query images; Right: top-5 retrieved satellite-view images, ranked from left (highest) to right (lowest). Correctly matched satellite images are highlighted with red bounding boxes.}
    \label{fig:CVUSA_res}
\end{figure*}

\noindent{\textbf{Visualization of view neural maps.}}
To qualitatively assess the behavior of the view-specific neural maps and enhance the interpretability of our method, we visualize the learned view neural maps at different feature levels in Fig.~\ref{fig:visualization}. Each sub-figure shows the ground-view neural map (left) and the corresponding satellite-view map (right), transformed via the proposed NEC $\Gamma_{bev}$. We observe that $\mathcal{N}_g$ primarily attends to sparse local regions at the shallow layers, while $N_s$ shows stronger activations across multiple locations. It indicates that the neural maps capture low-level spatial structures at earlier stages. As the network deepens, $N_g$, gradually expands its receptive field, allowing it to capture broader spatial relationships, while $N_s$ becomes more concentrated in specific central regions. This behavior aligns with the spatial distribution patterns observed in the CVUSA dataset, where key landmarks are typically centrally located. These results highlight the importance of explicitly modeling cross-view interactions to improve feature alignment in IRCVGL tasks.

\noindent{\textbf{Visualization of retrieval results.}}
Figure \ref{fig:CVUSA_res} illustrates qualitative retrieval examples obtained by our proposed CLNet on the CVUSA dataset. We show three sets of visualization results here, corresponding to (I) hit results, (II) top 5 results, and (III) unsuccessful retrieval results (i.e., no retrieval results in the top 5, the red box is the ground truth). 

As observed in Fig. \ref{fig:CVUSA_res} (I), our model successfully retrieves the correct satellite-view images within the top-5 predictions (highlighted with red boxes). Notably, the retrieved results exhibit strong visual and semantic correspondence to the query ground-view images, effectively capturing key spatial structures such as intersections, road curvature, and distinctive landmarks. Moreover, as shown in Fig. \ref{fig:CVUSA_res} (II), our method can correctly identify the corresponding satellite image in the top 5 retrieval results,  even in more challenging scenarios that contain multiple visually similar objects or have weak discriminative features. These experimental results demonstrate the strong robustness of the proposed method against ambiguous visual cues. We also present a set of failure examples in Fig. \ref{fig:CVUSA_res} (III).
It can be seen that when the reference image contains very little saliency information, our correspondence learning method fails to capture effective matching cues and find the correct matching images. In the future, we will explore more effective methods to overcome these extreme scenarios. This qualitative evaluation further confirms CLNet’s effectiveness in modeling cross-view correspondences and highlights its practical value in complex geo-localization scenarios. 

\section{Conclusion}
In this study, we propose \OURS to enhance IRCVGL by coupling the feature learning of ground and satellite views. Firstly, \OURS designs the ground-view neural maps to aggregate global contextual information. Secondly, the lightweight Neural bird's-Eye view Converter (NEC) bridges cross-view gaps by generating corresponding satellite-view neural maps according to the ground-view neural map. Finally, the Global Feature Recalibration (GFR) module injects correspondence priors into feature learning, enhancing cross-view feature consistency through neural map-guided feature enhancement. This work demonstrates the effectiveness of explicitly modeling latent correspondences for improving IRCVGL performance.

\bibliographystyle{plain}
\bibliography{neurips_2025}

\appendix

\section{Supplemental material}
\label{sec:suppl}

\subsection{Datasets}
\noindent{\textbf{CVUSA}~\cite{workman2015wide}} contains 35,532 pairs of images for training and 8,884 pairs of images for evaluation. The street view images are 224 x 1232, and the size of the satellite images is 750 x 750. By warping the ground-view images through the camera's extrinsic parameters, all ground-satellite image pairs are aligned, and all ground-view images are located in the center of the satellite images. 

\noindent{\textbf{CVACT}~\cite{liu2019lending}} is similar to CVUSA overall, except that it contains an additional test set, totaling 92,802 image pairs. In addition, most of the scenes in the dataset are urban from Australia, which can be considered as a different scene style compared with CVUSA dataset. At the same time, it has a larger image resolution. The size of the ground images is 832 x 1664, while the size of the satellite images is 1200 x 1200. The query images and reference images in CVUSA and CVACT have a one-to-one relationship in that each query corresponds to one reference, and the ground images and satellite images are also center-aligned.

\noindent{\textbf{VIGOR}~\cite{zhu2021vigor}} was collected from four cities in the United States, \textit{i.e.}, New York, Seattle, San Francisco, and Chicago. It contains two partitions, which are Same-area and Cross-area. The Same-area uses data from all cities for training and evaluation. The Cross-area uses New York and Seattle as the training set and San Francisco and Chicago for evaluation. Unlike the two datasets mentioned above, the positional relationship between the ground view image and the satellite image in this dataset contains two situations: positive and semi-positive. Concretely, each ground view image contains three semi-positive satellite images, which means the ground view image is located at the non-center region in three satellite images. Such a setting leads to a more difficult situation.

\noindent{\textbf{University-1652}~\cite{zheng2020university}} is proposed to bridge the gap between the drone-view and satellite-view. Different from the above three datasets that contain ground-view and satellite-view image pairs, the dataset includes multiple drone-view images for the same building. For instance, the authors collect 50, 218 images for training, and there are 54 drone-view images for each satellite-view image. For the benchmarking part, over 1, 652 buildings of 72 universities are captured to ensure the generalization of test results. Also, because not all satellite-view images have their reference drone-view images, we report the average precision (AP) for the dataset.

\subsection{Training settings}
In all our experiments, the batch size is 128 for CVUSA and CVACT, and 96 for VIGOR, We train the model on 8 $\times$ NVIDIA RTX 3090 GPUs and use 16 worker threads for PyTorch dataloader. We also employ random flip, random rotate, and ColorJitter for data augmentation. In addition, the optimizer is AdamW, with a learning rate of 0.001, and the other config is the default following PyTorch. We also set the training epoch to 40 and introduce cosine decay with a warmup as the learning rate schedule for better convergence.

\end{document}